\documentclass[10pt,twocolumn,letterpaper]{article}

\usepackage{iccv}
\usepackage{times}
\usepackage{epsfig}
\usepackage{graphicx}
\usepackage{amsmath}
\usepackage{amssymb}
\usepackage{caption}
\usepackage{subcaption}
\usepackage{bbm}
\usepackage{booktabs}
\usepackage[export]{adjustbox}

\usepackage[pagebackref=true,breaklinks=true,letterpaper=true,colorlinks,bookmarks=false]{hyperref}
\iccvfinalcopy 

\def\iccvPaperID{****} 

\ificcvfinal\fi

\def\P(#1){\Phelper#1|\relax\Pchoice(#1)}
\def\Phelper#1|#2\relax{\ifx\relax#2\relax\def\Pchoice{\Pone}\else\def\Pchoice{\Ptwo}\fi}
\def\Pone(#1){\Pr\left( #1 \right)}
\def\Ptwo(#1|#2){\Pr\left( #1 \mid #2 \right)}
\def\Pr{Pr}

\begin{document}

\title{SHARP Challenge 2023: Solving CAD History and pArameters Recovery from Point clouds and 3D scans. Overview, Datasets, Metrics, and Baselines.}


\author{Dimitrios Mallis$^{\star}$\\
{\tt\small dimitrios.mallis@uni.lu}
\and
Sk Aziz Ali$^{\star}$\\
{\tt\small skaziz.ali@uni.lu}
\and
Elona Dupont$^{\star}$\\
{\tt\small elona.dupont@uni.lu}
\and
Kseniya Cherenkova$^{\star \dagger}$\\
{\tt\small kseniya.cherenkova@uni.lu}
\and
Ahmet Serdar Karadeniz$^{\star}$\\
{\tt\small ahmet.karadeniz@uni.lu}
\and
Mohammad Sadil Khan$^{\star}$\\
{\tt\small mohammadsadil.khan@uni.lu}
\and
Anis Kacem$^{\star}$\\
{\tt\small anis.kacem@uni.lu}
\and
Gleb Gusev$^{\dagger}$\\
{\tt\small gleb@artec3d.com}
\and
Djamila Aouada$^{\star}$\\
{\tt\small  djamila.aouada@uni.lu}
\and
$^{\star}$SnT, University of Luxembourg\\
\and
$^{\dagger}$ Artec 3D
}

\maketitle
\ificcvfinal\thispagestyle{empty}\fi

\begin{abstract}
Recent breakthroughs in geometric Deep Learning (DL) and the availability of large Computer-Aided Design (CAD) datasets have advanced the research on learning CAD modeling processes and relating them to real objects. In this context, 3D reverse engineering of CAD models from 3D scans is considered to be one of the most sought-after goals for the CAD industry. However, recent efforts assume multiple simplifications limiting the applications in real-world settings. The SHARP Challenge 2023 aims at pushing the research a step closer to the real-world scenario of CAD reverse engineering 
through dedicated datasets and tracks. In this paper, we define the proposed SHARP 2023 tracks, describe the provided datasets, and propose a set of baseline methods along with suitable evaluation metrics to assess the performance of the track solutions. All proposed datasets\footnote{
\href{https://cvi2.uni.lu/cc3d}{https://cvi2.uni.lu/cc3d-data}} along with useful routines and the evaluation metrics\footnote{\href{https://gitlab.uni.lu/cvi2/iccv2023-sharp-challenge}{https://gitlab.uni.lu/cvi2/iccv2023-sharp-challenge}
} are publicly available. 
\end{abstract}

\vspace{-0.3cm}
\section{Introduction}
\begin{figure}[!ht]
\setlength{\belowcaptionskip}{-0.5cm}
\begin{center}

   \includegraphics[width=1.0\linewidth]{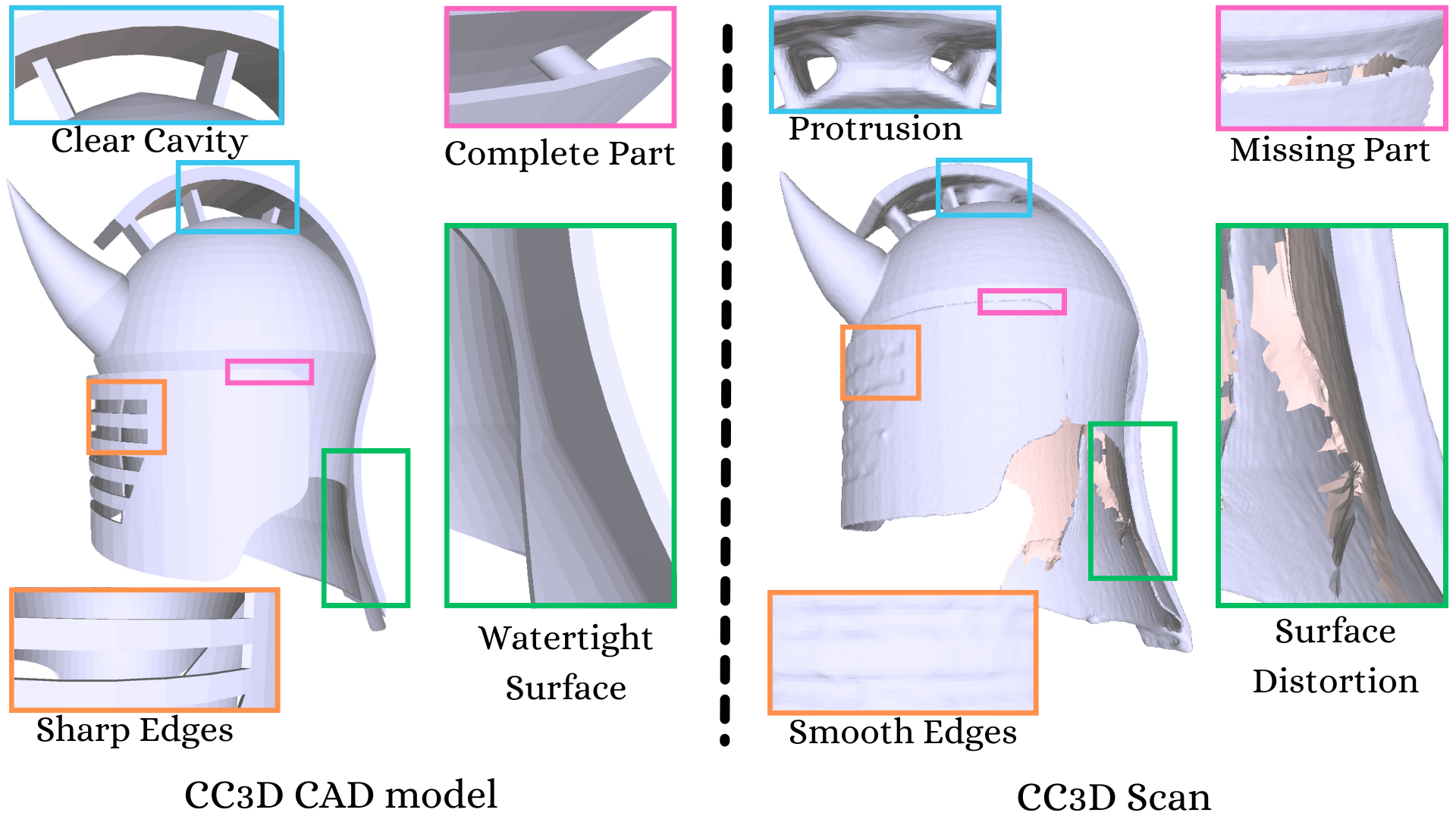}
\end{center}
\vspace{-0.6cm}
\caption{3D scans in CC3D dataset~\cite{cherenkova2020pvdeconv} contain various artefacts (unwanted protrusions, smoothness over surfaces/edges, missing regions). Tackling these artefacts is essential for robust Scan-to-CAD algorithms.
}
\label{fig:CC3DDatasetStat_Quality}
\end{figure}

\textit{3D reverse engineering is defined as the deduction of intermediate design steps, complete history, and final intent in a reasonable fashion from a given 3D scan of its corresponding Computer-Aided Design (CAD) model}.
%
%
%
%

In today’s digital era, using CAD software is the standard approach for designing objects ahead of manufacturing. However, CAD modeling cannot be seen as straightforward and simple procedural design, as it requires the skills of highly qualified engineers. 
Consequently, 3D reverse engineering has been a long-sought-after goal in CAD industry due to the huge resources and time that it could save~\cite{uy2022point2cyl,dupont2022cadops}. Such technique, also referred to as \textit{Scan-to-CAD}, consists of scanning objects and automatically concluding their corresponding CAD models. Recently, solving this problem has attracted a large interest from the Computer Vision and Graphics research communities~\cite{wu2021deepcad,koch2019abc,li2022free2cad,matveev2022def,lambourne2021brepnet,dupont2022cadops,uy2022point2cyl,willis2021fusion,izadinia2017im2cad,Li:2020:Sketch2CAD}, thanks to the huge advances in 3D geometric deep learning and the availability of open repositories for CAD models. The idea is to learn a mapping from 3D scans to CAD models using the available models submitted by the designers in open repositories such as OnShape~\cite{onshape} and 3D Content Central~\cite{3DContentCentral}. While the representation of 3D scans is well established and often consists of meshes or point clouds, CAD model representations may vary 
depending on the use case. 

One approach for representing CAD models is through its final shape as
a collection of geometric primitives, e.g., cylinders, cubes. 
Such representation is expressed by Constructive Solid Geometry (CSG)  modeling, however, modern CAD workflows use Feature-based modeling as a superior alternative. 
%
%
%
%
Feature-based modeling is 
widely used as it allows to create solids by iteratively adding features such as holes, slots, or bosses, thus giving more expressiveness to designers~\cite{xu2021inferring}. In this setting, the final object's geometry and topology are stored as a \textit{Boundary Representation} (B-Rep) which is a graph structure encoding parametric faces and edges, loops, and vertices~\cite{lambourne2021brepnet}. Accordingly, recent works have tried to infer some of the attributes of B-Reps from point clouds to enable their editability. Some of them focused on inferring the edges~\cite{cherenkova2023sepicnet,zhu2023nerve,matveev2022def}, other attempts considered the prediction of the faces~\cite{sharma2020parsenet,li2019supervised}, and few of them aimed at predicting both faces and edges~\cite{li2023surface,guo2022complexgen}.

CAD modeling can also be seen as the process that allows the creation of the final model referred to as \textit{design history}. Design history consists of the set of ordered steps that were followed by the designer using a CAD software. In feature-based modeling, these ordered steps involve the drawing of CAD sketches~\cite{Li:2020:Sketch2CAD,seff2021vitruvion,willis2021engineering} followed by CAD operations such as extrusion, revolution, etc~\cite{wu2021deepcad,willis2021fusion}. Thanks to the availability of dedicated datasets~\cite{koch2019abc,willis2021fusion}, multiple works in literature focused on learning this design history in order to automatically generate plausible CAD models~\cite{wu2021deepcad,xu2022skexgen}, complete partial designs according to the intent of designers~\cite{xu2023hierarchical}, or predict it from point clouds~\cite{uy2022point2cyl,wu2021deepcad,lambourne2022reconstructing,li2023secad}. 



From the two aforementioned representations for CAD models, the problem of Scan-to-CAD can be seen as either related to recovering some attributes of the B-Rep from the corresponding 3D scan, or inferring the design history that allowed its creation.  
Despite recent findings, this problem is far from being solved. In particular, the current efforts remain very limited in the context of real-world scenarios due to the strong assumptions that are made to over-simplify the problem. For instance, it is very common to consider simple objects (e.g., cubes and cylinders) and restrict the study to the basic extrusion operation~\cite{wu2021deepcad,xu2022skexgen,uy2022point2cyl}. Furthermore, most of the works in literature assimilate 3D scans to sampled point clouds on CAD models~\cite{wu2021deepcad,uy2022point2cyl,lambourne2022reconstructing} which is not the case in real world scenarios. Indeed, as mentioned in~\cite{cherenkova2023sepicnet,cherenkova2020pvdeconv}, 3D scans are often subject to scanning artifacts resulting in smoothed high-level geometrical details and missing parts. Compared to uniformly sampled point clouds on CAD models, these artifacts make the problem of Scan-to-CAD more challenging.

The aim of the SHARP challenge 2023 is to encourage and help the research community to get a step closer to the real-world setting of inferring CAD history and parameters of objects from their 3D scans. In particular, different variants of the CC3D dataset~\cite{cherenkova2020pvdeconv} are proposed along with three different tracks. It is important to highlight that the CC3D dataset has the advantage of bringing pairs of realistic 3D scans with their corresponding CAD models, thus enabling a more realistic scenario of Scan-to-CAD as compared to using sampled point clouds on CAD models. Furthermore, as stated in~\cite{dupont2022cadops}, the CAD models in CC3D dataset are more complex in nature than the ones used in literature~\cite{wu2021deepcad,willis2021fusion}. The tracks proposed in SHARP challenge span over the design history and the B-Rep of the CAD models, with one of them tackling the inference of B-Rep parametric edges from 3D scans, the second focusing on the per-point segmentation of B-Rep faces from scans, and the third aiming at segmenting scans into ordered CAD operation steps and types of the corresponding design history. Simple baseline solutions to the aforementioned tracks are also proposed along with a set of dedicated metrics to assess their performance. 

The rest of the paper is organised as follows. Section~\ref{sect:data-tracks} defines the different tracks introduced in the SHARP challenge and describes the datasets. In Section~\ref{sect:baselines}, the baseline methods for the proposed tracks are described. The evaluation metrics used to assess the performance of the methods are described in Section~\ref{sect:metrics}. Section~\ref{sect:results} reports the results of the baseline methods. Finally, conclusions and perspectives of the proposed challenge are drawn in Section~\ref{sec:conclusions}.







\section{Challenge and Dataset Description}
\label{sect:data-tracks}
\begin{figure*}[t]
\begin{center}
\setlength{\belowcaptionskip}{-0.65cm}
\includegraphics[width=\linewidth]{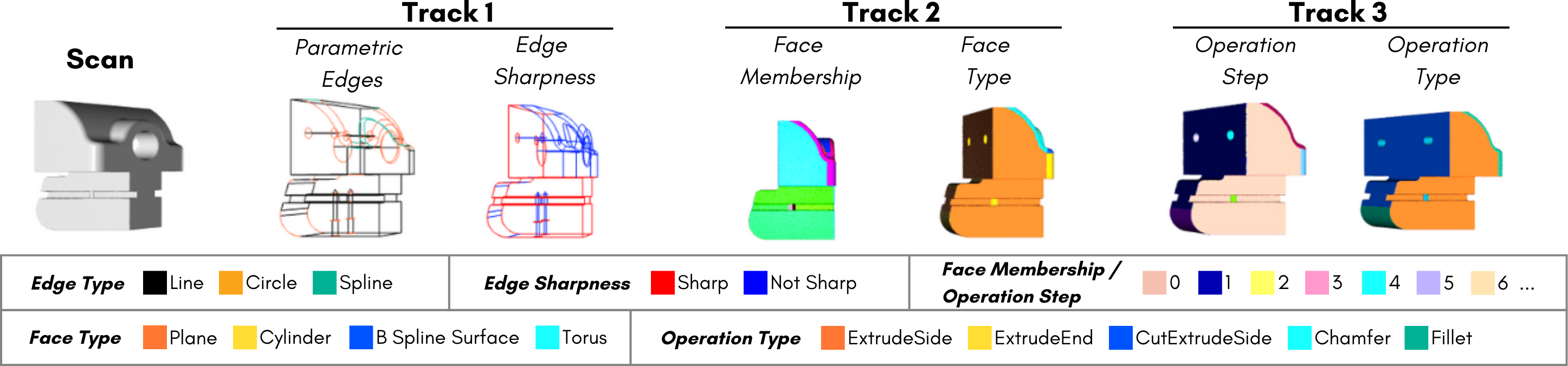} 
\vspace{-0.6cm}
\caption{Predictions targets for the SHARP Challenge 2023. Proposed tracks relate to recovering geometrical and topological properties of the B-rep \textit{(track 1 and track 2)}, as well as attributes of the design history of the CAD model \textit{(track 3)}.}
\label{fig:track_description}
\end{center}
\end{figure*}
The SHARP 2023 challenge focuses on three different tasks to bridge the gap between 3D scans and their corresponding CAD models. Three  versions of the CC3D dataset~\cite{cherenkova2020pvdeconv} are used in these tracks. The CC3D dataset is derived from open CAD repositories such as 3D Content Central~\cite{3DContentCentral}. Unlike other alternatives such as ABC
dataset~\cite{koch2019abc}, where the noise is usually synthetically added to the sampled point cloud, the 3D scans were obtained by virtually scanning the corresponding CAD models, using a proprietary 3D scanning pipeline developed by Artec3D~\cite{artec3dProfessionalScanners}. As shown in Figure~\ref{fig:CC3DDatasetStat_Quality}, the 3D shapes in CC3D dataset may have artifacts in the form of missing parts or protrusions due to specifics of a scanning system. The total number of samples of the CC3D dataset used in SHARP challenge is $31185$, split into $25612$ training samples and $5573$ test samples. Note that the same sets are used for all three tracks and the corresponding B-Rep models of the training set are also provided. The overall objective is to infer different information for the CAD model given a 3D scan. While \textit{Track 1} and \textit{Track 2} focus on inferring geometrical and topological properties of the Boundary Representation of the CAD model, \textit{Track 3} is centered around predicting attributes of the design history of the CAD model. More formally, let us consider a 3D scan represented by a point cloud $\mathbf{X}~=~[\textbf{x}_{1}, \textbf{x}_{2}, \ldots, \textbf{x}_{N}] \in \mathbb{R}^{N \times 3}$, where $\textbf{x}_{i}$ denotes the 3D coordinates of a point $i$ and $N$ the number of points. The objective of \textit{Track~1} and \textit{Track~2} is to predict different attributes from the B-Rep $\mathcal{B}$ of the corresponding CAD model, while \textit{Track 3} aims at recovering other attributes from the design history $\mathcal{H}$ of the CAD model. The datasets and the attributes of each track are described next. 

\vspace{0.2cm}
\textbf{Track 1: Parametric Sharp Edge Inference.}
\label{sec:track1}
Given a 3D scan $\mathbf{X}$, the goal of \textit{Track 1} (see Figure~\ref{fig:track_description}) is to recover: (1) the set of parametric edges $\{\textbf{e}_{j}\}_{j=1}^{N_e} \in \mathcal{E}$ that are present in the the B-Rep $\mathcal{B}$, where $\mathcal{E}$ denotes the set of 3D parametric curves among circles, lines, and splines; (2)~and a sharpness label $s_{j} \in \{0,1\}$ indicating whether a recovered edge $\textbf{e}_{j}$ is sharp ($s_{j}=1$) or not ($s_{j}=0$). Note that recovering these parametric B-Rep edges and their sharpness from 3D scans is critical for CAD reverse engineering. Indeed, these edges encode the topology and the geometry of the boundary of the B-Rep and some of them can be part of the sketches drawn by the designer. 


 
\begin{figure}
\setlength{\belowcaptionskip}{-0.47cm}
     \centering
     \begin{subfigure}[b]{0.68\linewidth}
         \centering
         \includegraphics[width=\linewidth]{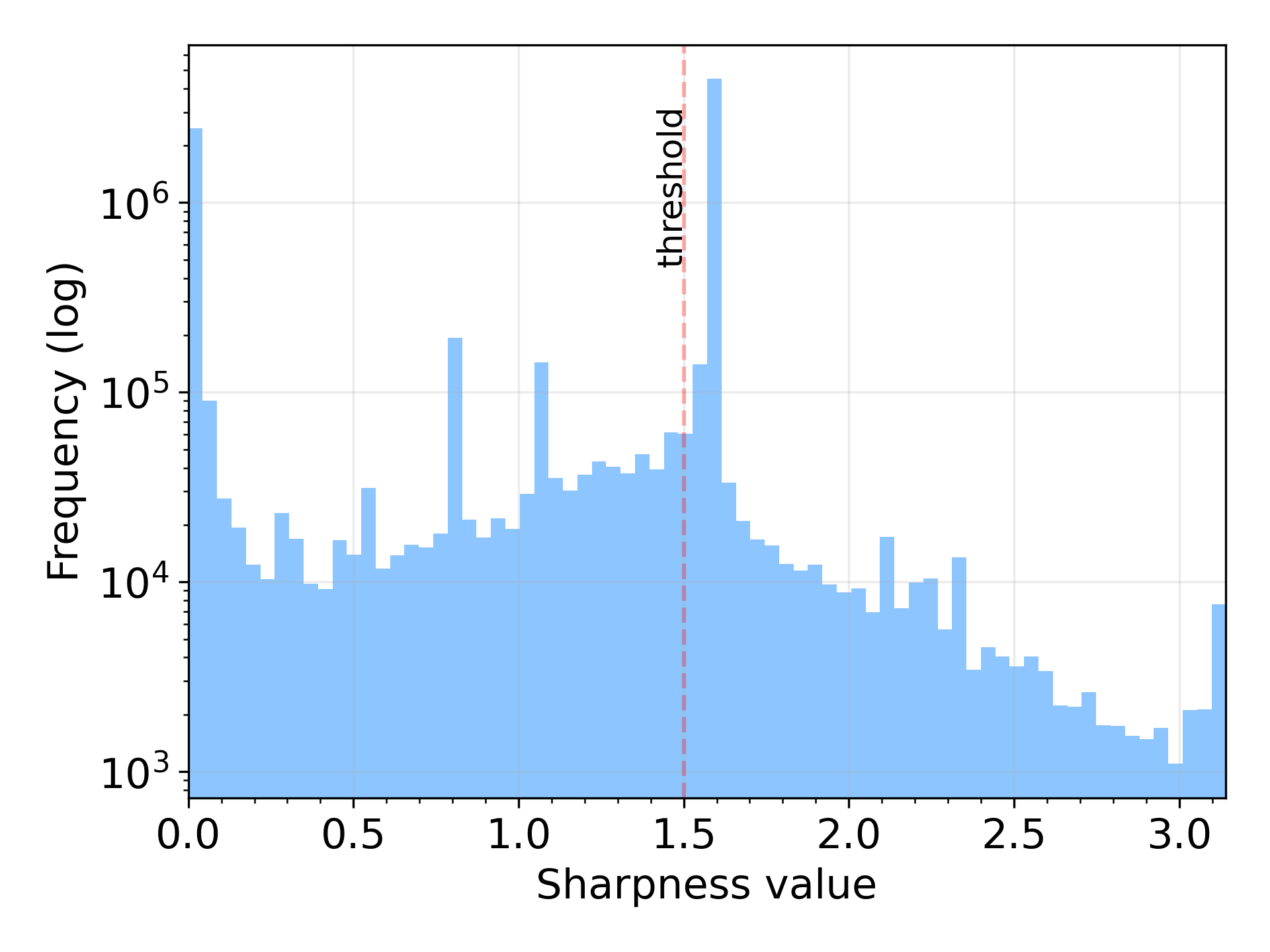}
         \label{fig:sharpness_hist}
     \end{subfigure}
     \begin{subfigure}[b]{0.3\linewidth}
         \centering
         \includegraphics[width=\linewidth]{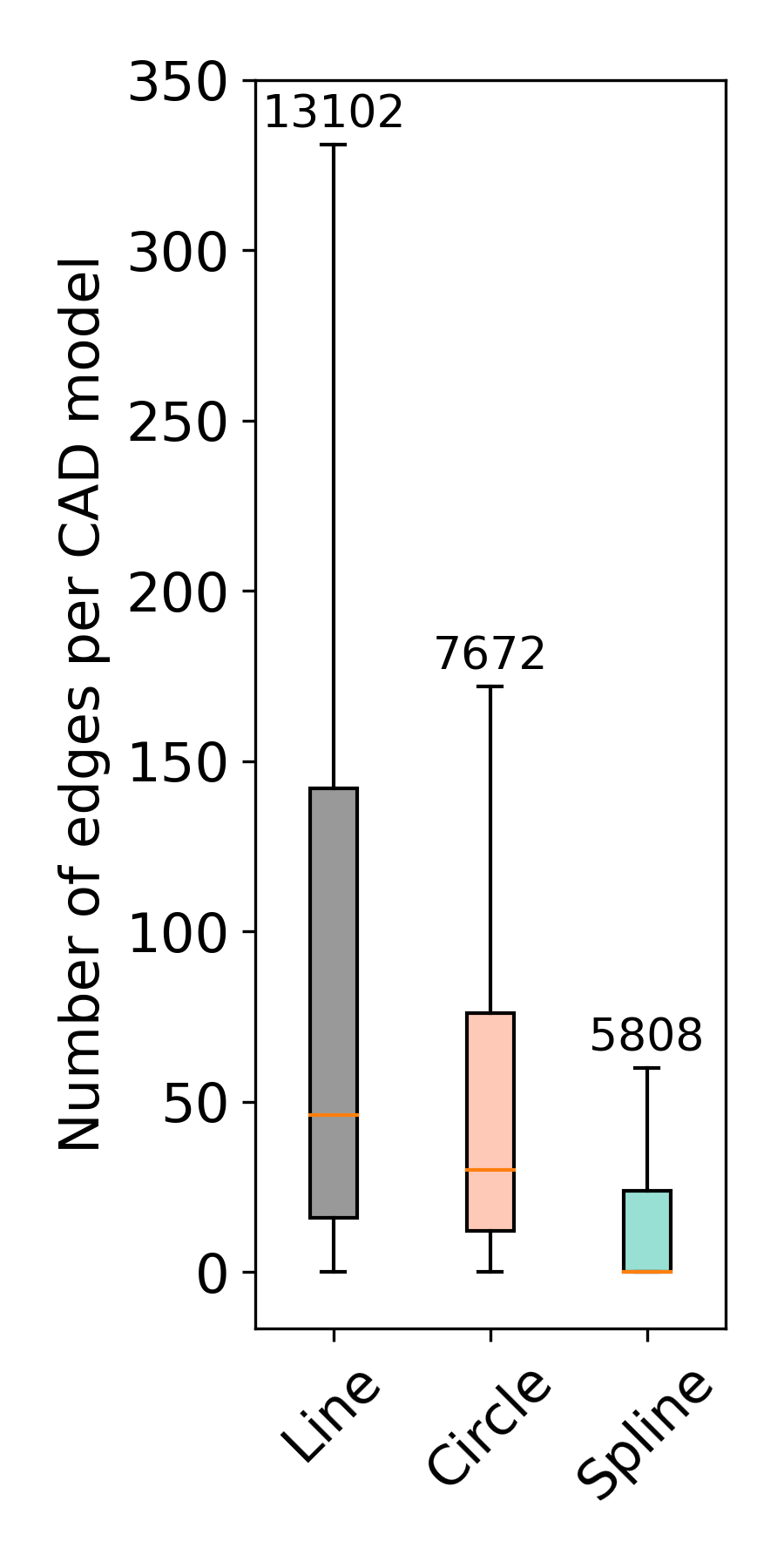}
         \label{fig:edge_type_boxplot}
     \end{subfigure}
     \vspace{-0.6cm}
        \caption{\textit{(left)} Histogram of sharpness values for the edges in the CC3D-PSE dataset. \textit{(right)} Boxplot of the number of edges per CAD models for different edge types.}
        \label{fig:track 1 stat graphs}
\end{figure}

 \textit{CC3D-PSE dataset:} This dataset consists of a set of 3D scans, annotated with a set of parametric edges and  corresponding sharpness values. Both the ground truth edges and the sharpness are extracted from the B-Rep of the corresponding CAD model.  The parametric edges are directly extracted from the B-Reps using OpenCascade API~\cite{opencascadeIntroductionOpen}, and their sharpness value is computed as the angle between the normals of the two neighboring surface patches to the edge. The distribution of the sharpness values (\textit{left} of Figure~~\ref{fig:track 1 stat graphs}) reveal that about $30\%$ of the edges have a sharp value lower than a corresponding angle of 10 degrees. Also, about $50\%$ of the edges have a sharpness of $1.57$ corresponding to an angle of $90$ degrees, which is expected in the context of CAD models. Three types of parametric edges are considered (lines, circles and splines). From the distribution of the different edge types per model (\textit{right} of Figure~\ref{fig:track 1 stat graphs}), we observe that the line type is the most common type. Additionally, about $15\%$ of the CAD models in the CC3D-PSE have more than 500 edges, demonstrating the complexity of the dataset.
 The annotations of different edges types are parametrized as follows: (1) A line is parameterized by a start and an end point \hbox{$\textbf{p}_s=(p_s^x, p_s^y, p_s^z)\in\mathbb{R}^3$ and $\textbf{p}_e=(p_e^x, p_e^y, p_e^z)\in \mathbb{R}^3$}. (2) A circle (or circular arc) is defined by a start point \hbox{$\textbf{p}_s=(p_s^x, p_s^y, p_s^z)\in\mathbb{R}^3$}, and an end point \hbox{$\textbf{p}_e=(p_e^x, p_e^y, p_e^z)\in\mathbb{R}^3$}, a center point \hbox{$\textbf{p}_c=(p_c^x, p_c^y, p_c^z)\in \mathbb{R}^3$}, a normal vector \hbox{$\Vec{\mathbf{n}}=(n_x,n_y,n_z)\in \mathbb{R}^3$,} and \hbox{a radius $r\in\mathbb{R}$}. (3) A spline is parameterized by a degree $K\in\mathbb N$ and a set of keypoints $\textbf{P}_k=[\textbf{p}_k^1, \textbf{p}_k^2, \dots, \textbf{p}_k^K]$, where each $\textbf{p}_k=(p_k^x, p_k^y, p_k^z)\in\mathbb{R}^3$. Note that the original CC3D-PSE dataset has been used in the SHARP 2022 challenge~\cite{sharp22c2} and in~\cite{cherenkova2023sepicnet} for parametric sharp edge inference. This updated version includes all B-Rep edges and their sharpness. 

\vspace{0.2cm}
\textbf{Track 2: Boundary-Representation (B-Rep) Face Segmentation}.
\label{sec:track2}
Given a 3D scan $\mathbf{X}~=~[\textbf{x}_{1}, \textbf{x}_{2}, \ldots, \textbf{x}_{N}]$, the goal of \textit{Track 2} (see Figure~\ref{fig:track_description}) is to simultaneously segment it into: \textbf{(1)} B-Rep face memberships, which consists of predicting for each point $\textbf{x}_{i}$ a face that it should belong to in the B-Rep $\mathcal{B}$; \textbf{(2)} the type of that B-Rep face. In other words, the objective is to predict a point-to-face membership matrix $\textbf{M} \in \{0,1\}^{N \times N_f}$, where $N_f$ denotes the number of the faces in the B-Rep $\mathcal{B}$, and a per-point face type matrix $\textbf{T} \in \{0,1\}^{N \times N_{f_t}}$, where $N_{f_t}$ is the number of B-Rep face types considered. Note that each point $\textbf{x}_{i}$ can only belong to a single face and should have a unique type, which suggests that each row of $\textbf{M}$ (and $\textbf{T}$) should have a single entry with $1$ and $0$ anywhere else. 
 The objective of this track is to infer the B-Rep face structure from raw 3D scans which is extremely important for CAD reverse engineering.



\begin{figure}
\setlength{\belowcaptionskip}{-0.47cm}
     \centering
     \begin{subfigure}[b]{0.6\linewidth}
         \centering
         \includegraphics[width=\linewidth]{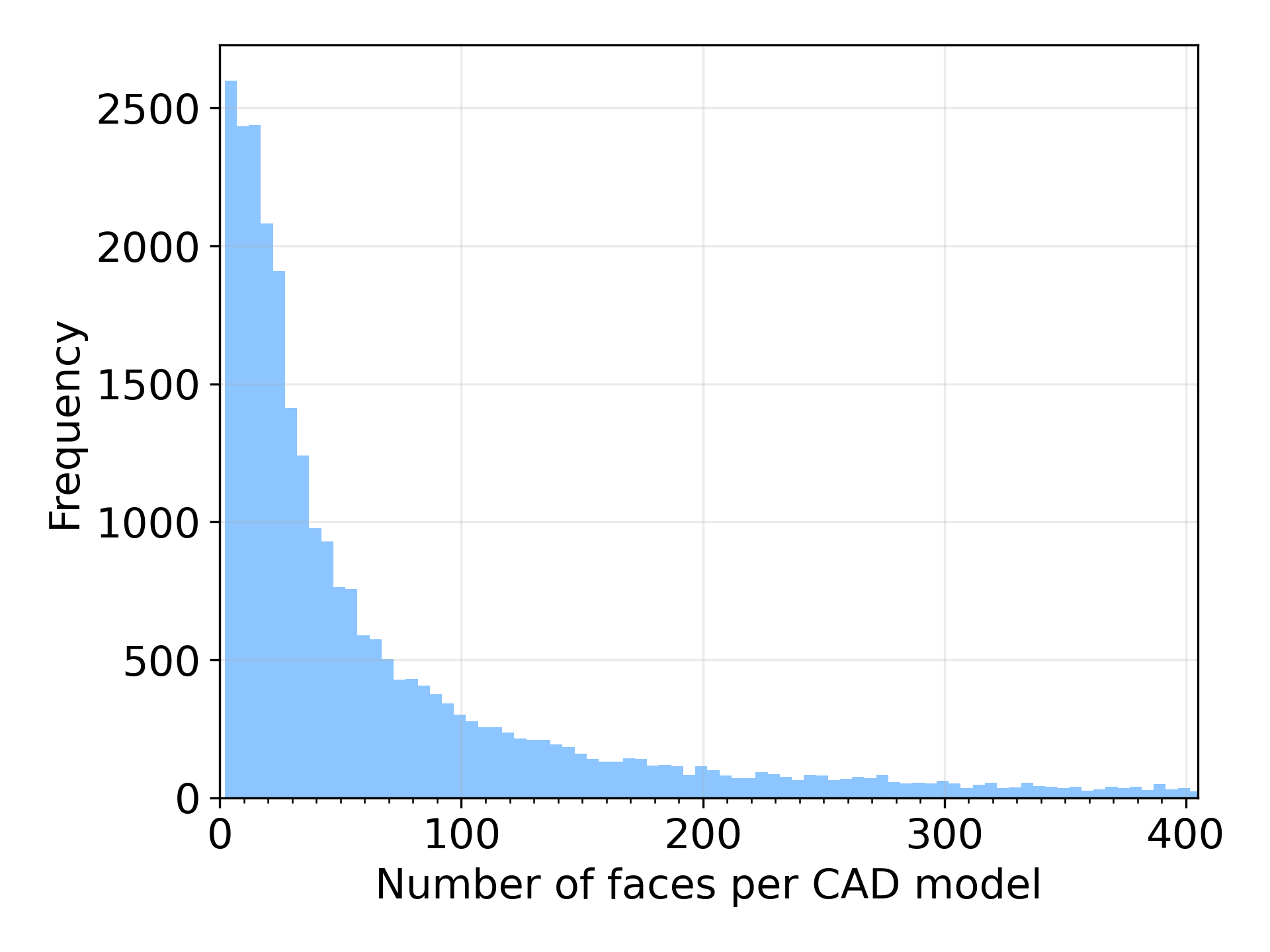}
         \label{fig:num_faces_hist}
     \end{subfigure}
     \begin{subfigure}[b]{0.38\linewidth}
         \centering
         \includegraphics[width=\linewidth]{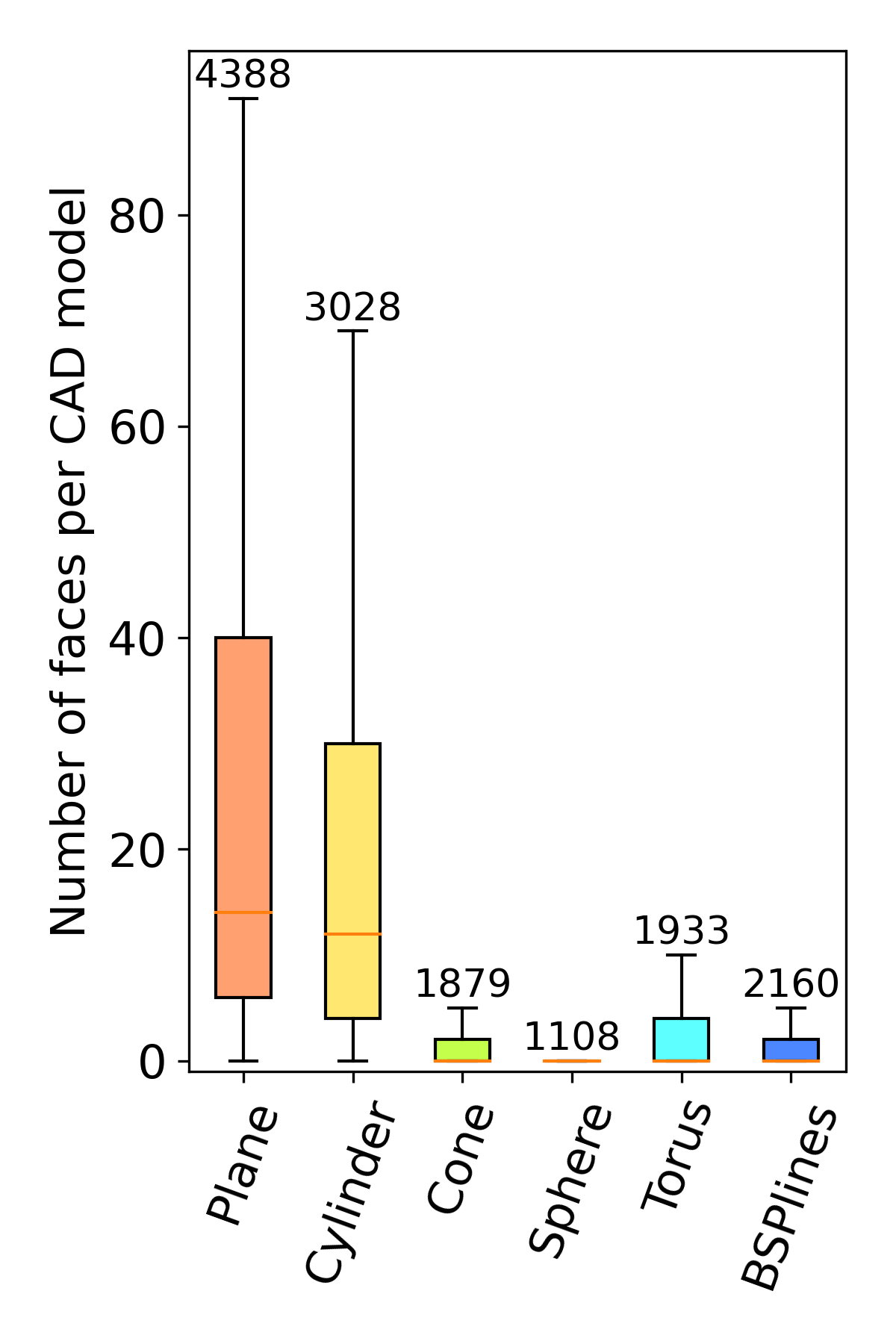}
         \label{fig:face_type_boxplot}
     \end{subfigure}
     \vspace{-0.6cm}
        \caption{\textit{(left)}: Histogram of the number of faces per CAD model for $95\%$ the CC3D-BRepFace dataset. \textit{(right)} Boxplot of the number of faces per CAD models for the different types of faces per CAD model.}
        \label{fig:track 2 stat graphs}
\end{figure}

\textit{CC3D-BRepFace dataset:} \textit{Track 2} uses C3D-BRepFace, a newly introduced version of the CC3D dataset~\cite{cherenkova2020pvdeconv} that brings the annotations of the B-Rep face structure to 3D scans. In particular, the 3D scans were annotated with the face membership and type labels according to the corresponding B-Rep. This is done by processing the B-Reps with OpenCascadeAPI~\cite{opencascadeIntroductionOpen} and transferring the labels to the points of the corresponding 3D scans through nearest neighbor assignment. This results in two annotations per point for each 3D scan: \textbf{(1)} an ID of the B-Rep face that it belongs to and; \textbf{(2)} the type of the face among six possible surface types (Plane, Cylinder, Cone, Sphere, Torus, or B-Spline). Figure~\ref{fig:track 2 stat graphs} shows statistics of the CC3D-BRepFace dataset. 
The CC3D-BRepFace dataset contains a wide range of model complexities with $50\%$ of the models containing more that $37$ faces and the maximum number of faces per model being $4388$. 
We see that Plane surface is the most common face type in the dataset followed by Cylinder. Moreover, we found out that $\approx60\%$ of the models contain at least 3 different types of face types.
%
%
%
%
%
%
%
%
%

\vspace{0.2cm}
\textbf{Track 3: Operation Type and Step Segmentation}.
\label{sec:track3} Given a 3D scan $\mathbf{X}~=~[\textbf{x}_{1}, \textbf{x}_{2}, \ldots, \textbf{x}_{N}]$, the goal of \textit{Track 3} is to simultaneously segment into: (1) the ordered CAD operation steps that allowed its creation; (2)~the corresponding CAD operation types that were used by the designer. In other words, the objective is to predict a point-to-step membership matrix $\textbf{S}~\in~\{0,1\}^{N \times N_s}$, where $N_s$ denotes the number of steps used in the CAD history $\mathcal{H}$, and a per-point CAD operation type matrix $\textbf{O}~\in~\{0,1\}^{N \times N_{O_t}}$, where $N_{O_t}$ is the number of CAD operation types considered. Similarly to \textit{Track 2}, each point $\textbf{x}_{i}$ can only belong to a single CAD operation step and should have a unique CAD type, which suggests that each row of $\textbf{S}$ (and $\textbf{O}$) should have a single entry with $1$ and $0$ anywhere else. Although \textit{Track 3} and \textit{Track 2} seem to be conceptually similar, there are two main differences. Firstly, the membership task in \textit{Track 3} aims at inferring ordered CAD steps in contrast to \textit{Track 2} where the order of face membership does not matter. Secondly and most importantly, the nature of the labels that are targeted in the two tracks is totally different. While \textit{Track 2} focuses on the B-Rep face structure, \textit{Track 3} goes beyond B-Reps and aims at recovering the CAD operation history out of 3D scans. As highlighted in~\cite{dupont2022cadops,lambourne2021brepnet,wu2021deepcad}, recovering these operation types and steps is crucial for CAD reverse engineering as it not only provides information about how the model was constructed but also at which stage of the design the geometry was created.

\begin{figure}
\setlength{\belowcaptionskip}{-0.47cm}
     \centering
     \begin{subfigure}[b]{0.49\linewidth}
         \centering
         \includegraphics[width=\linewidth]{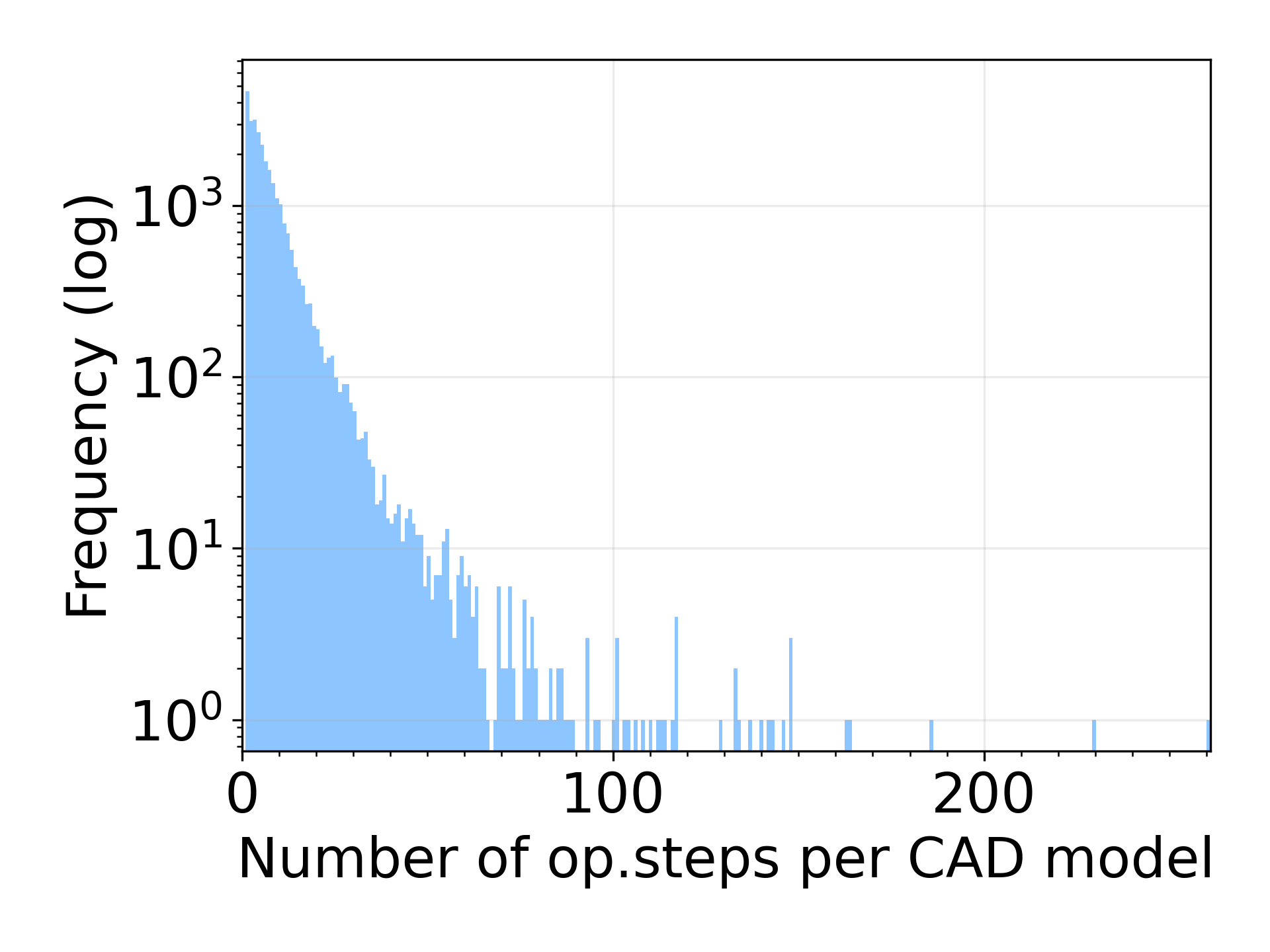}
         \label{fig:op_step_hist}
     \end{subfigure}
     \begin{subfigure}[b]{0.49\linewidth}
         \centering
         \includegraphics[width=\linewidth]{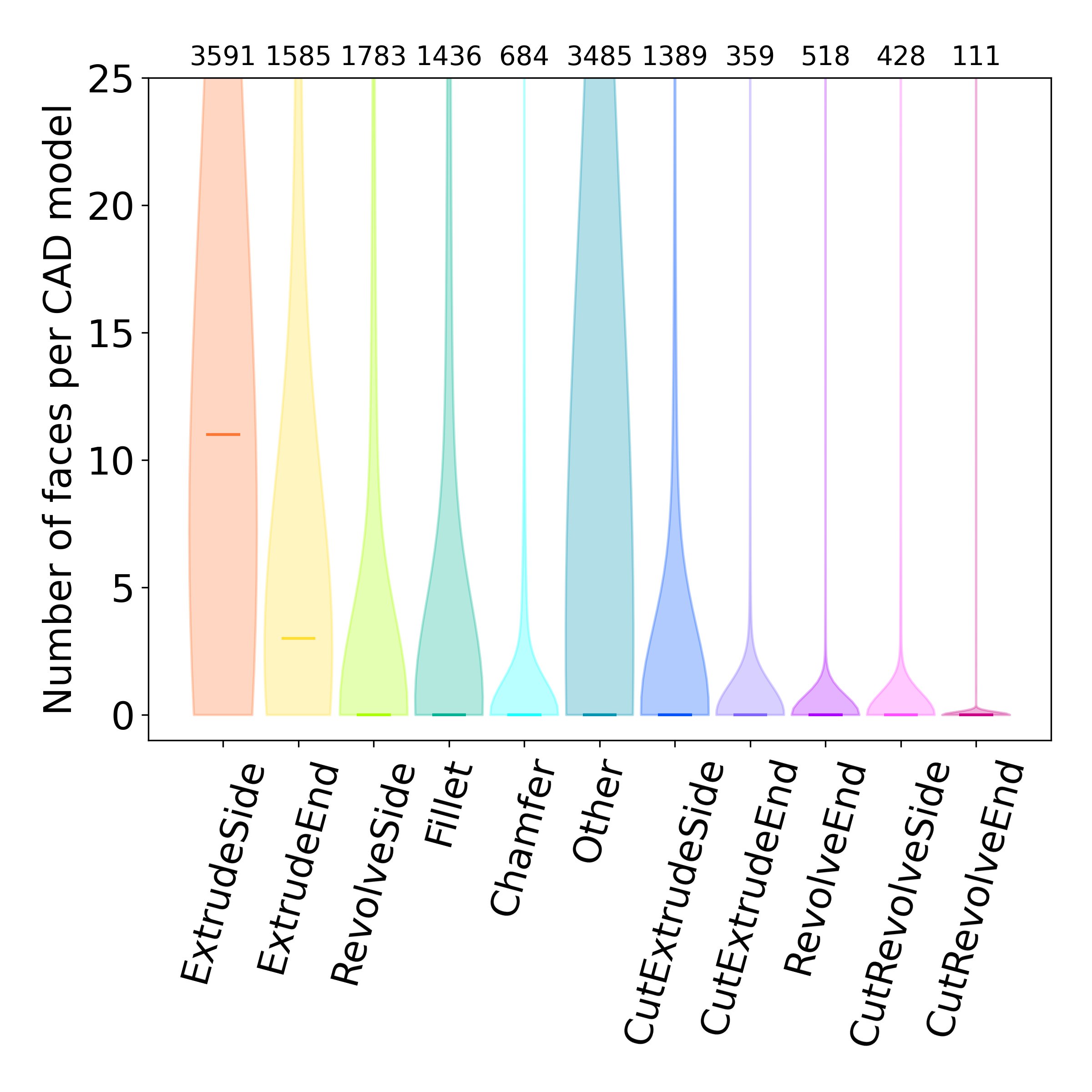}
         \label{fig:op_type_violinplot}
     \end{subfigure}
     \vspace{-0.6cm}
        \caption{\textit{(left)}: Histogram of the number of operation steps per CAD model in the CC3D-Ops  dataset. \textit{(right)}: violin plot of the number of faces per CAD models for the different operation types per CAD model. The horizontal lines represent the median values.}
        \label{fig:track 3 stat graphs}
\end{figure}

\begin{figure*}[t]
\begin{center}
\setlength{\belowcaptionskip}{-0.65cm}
\includegraphics[width=0.98\linewidth]{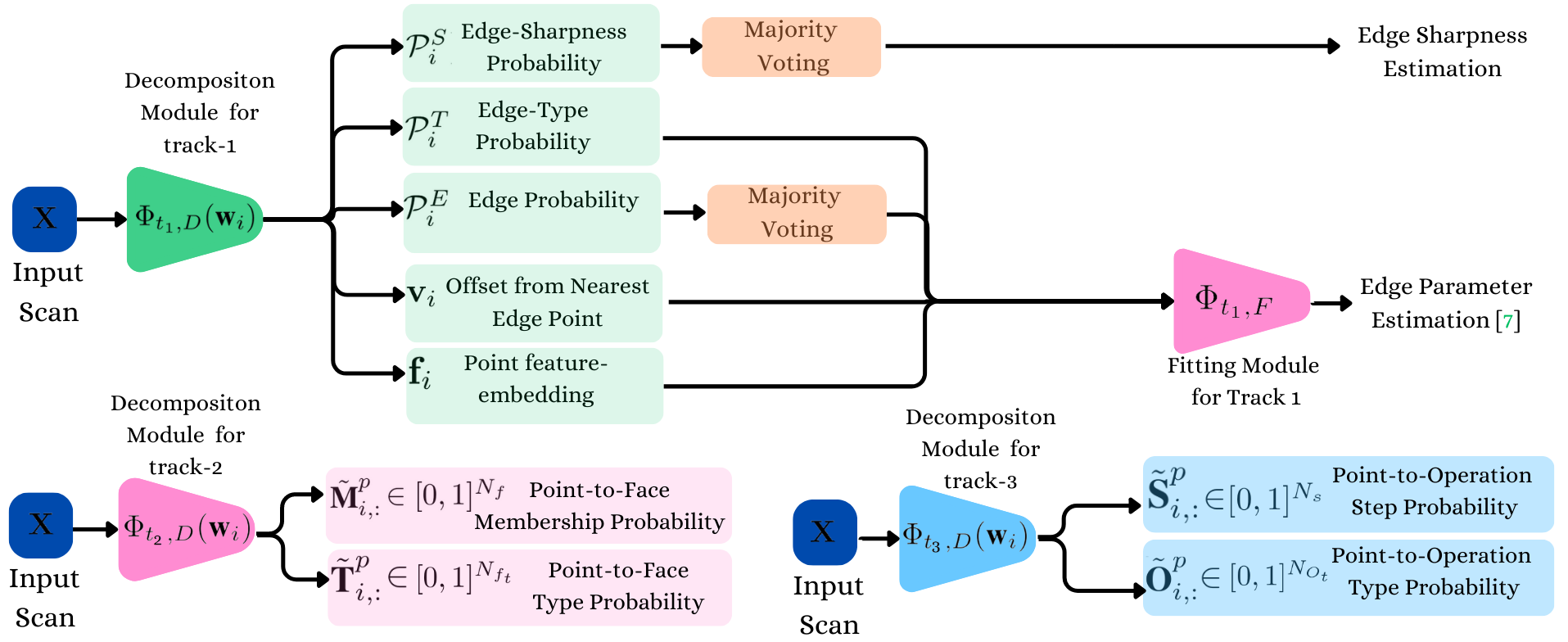} 
\vspace{-0.1cm}
\caption{Baseline model architecture for the three tracks. The input point cloud is first encoded through a PCVNN encoder. For \textit{Track 1}, the fitting module of \cite{Cherenkova2023SepicNetSE} is used to produce the set of parametric edges. For \textit{Track 2} and \textit{Track 3}, the mapping from the vertex embeddings to the target labels is learned directly through separate MLPs.
}
\label{fig:baseline_model}
\end{center}
\end{figure*}

\textit{CC3D-Ops dataset:} The dataset used in \textit{Track 3} extends the CC3D-Ops dataset introduced in~\cite{dupont2022cadops}. While the original CC3D-Ops dataset contains CAD operation type and step annotations on B-Rep faces, the extended version offers the same annotations on the corresponding 3D scans at the point level. In particular, each point of a 3D scan is labelled with: \textbf{(1)} a CAD operation step identifier and; \textbf{(2)}~a CAD operation type that can be one of the following types: ExtrudeSide, ExtrudeEnd, RevolveSide, Fillet, Chamfer, CutExtrudeSide, CutExtrudeEnd, RevolveEnd, CutRevolveSide, CutRevolveEnd, and Other. Figure~\ref{fig:track 3 stat graphs} shows statistics about these annotations. In the left part of this Figure, it can be observed that most of the models of the CC3D-Ops dataset were constructed in $50$ operation steps or less. In particular, $90\%$ of the models were created in $16$ or less operation steps, while the maximum number of steps per model is $261$. In the right part of Figure~\ref{fig:track 3 stat graphs}, the operation type distribution is shown in violin plot. It can be seen that the most common operation type is the extrusion type but the dataset also contains models with a large number of faces created from less common operations such as revolution. 





\section{Baseline Methods }
\label{sect:baselines}


We provide simple baselines for all three tracks introduced for the SHARP challenge. Across tracks, the innput 3D scan is represented as a point cloud $\mathbf{X}~=~[\textbf{x}_{1}, \textbf{x}_{2}, \ldots, \textbf{x}_{N}]$ that is uniformly downsampled to fixed number of points $N=10k$. For all three tracks, the input $\mathbf{X}$ is mapped to a learned per-point representation $\mathbf{W}~=~[\textbf{w}_{1}, \textbf{w}_{2}, \ldots, \textbf{w}_{N}] \in \mathbb{R}^{N \times 960}$ through a point cloud backbone $\Phi_b$. In this work, we model $\Phi_b$ as the Point-Voxel CNN introduced in \cite{liu2019pvcnn}. Then, $\mathbf{W}$ is processed by output heads $\Phi_{t_1}$, $\Phi_{t_2}$ and $\Phi_{t_3}$, developed to address tracks one to three as discussed next.

\vspace{0.2cm}
\textbf{Track 1 Baseline}. The goal of this track is to learn the mapping $\Phi_{t_1}$ from the learned representation $\mathbf{W}$ to a set of parametric edges $\{\textbf{e}_{j}\}_{j=1}^{N_e} \in \mathcal{E}$ and their corresponding sharpness labels $\{{s}_{j}\}_{j=1}^{N_e} \in \{0,1\}$. Our solution to this problem is based on the recently proposed SepicNet~\cite{Cherenkova2023SepicNetSE}. $\Phi_{t_1}$ comprises two modules, the \textit{decomposition module} $\Phi_{t_1,D}$ followed by the \textit{fitting module} $\Phi_{t_1,F}$ that are trained in an end-to-end manner (see  Figure~\ref{fig:baseline_model} (\textit{left})).

\textit{Decomposition module} $\Phi_{t_1,D}$: The decomposition module detects edge points, consolidates them along the edge and groups them into different segments with primitive types identified. More specifically, for each point representation $\textbf{w}_i$, we have $\Phi_{t_1,D}(\textbf{w}_i)~=~\{ \mathcal{P}^E_i, \mathbf{v}_i, \mathcal{P}_i^T, \mathbf{f}_i, \mathcal{P}_i^S\}$, where $\mathcal{P}^E_i~\in~[0,1]$ is the probability that the $i$'th point is an edge point, $\mathbf{v}_i~\in~\mathbb{R}^3$ is a displacement vector to the closest edge, $\mathcal{P}^T_i~\in~[0,1]^3$ denote the probabilities of the primitive type of the closest edge among three possible types, namely, line, circle, and spline, and $\mathbf{f}_i~\in~\mathbb{R}^{D_{emb}}$ is a per-point embedding that is used to cluster edge points into distinct segments. Finally, $\mathcal{P}^S_i~\in~[0,1]$ predicts the probability that the closest edge is sharp. 
To group points corresponding to the same segment, a differentiable version of the mean-shift clustering algorithm is used~\cite{Sharma2020ParSeNetAP}. Curve type and sharpness for each segment are determined through majority voting. Training labels for edge points, types, sharpness and displacement vectors are derived from the ground truth whereas the per-point embeddings are trained through a triplet loss that contrasts points from the same segment to points from different segments.

\textit{Fitting module} $\Phi_{t_1,F}$: The fitting module performs parameter estimation for each detected curve/segment. Curve parameters are estimated through least-squares fitting using differentiable SVD \cite{Ionescu2015TrainingDN} on the set of segmented edge points. Different formulations for differentiable curve parameter estimation are provided for lines, arcs, and splines. Thus, a fitting loss can be formed as a sum of distances between the points sampled on the predicted parameterized segments and points sampled uniformly on the ground truth segments. During training, the decomposition module is first pretrained for $100$ epochs and then jointly finetuned with the fitting module using a combination of decomposition and fitting losses. For a more thorough description of the fitting module, readers are referred to~\cite{Cherenkova2023SepicNetSE}. 

\vspace{0.2cm}
\textbf{Track 2 Baseline.} \label{track2_baseline}
The goal of this track is to predict a point-to-face membership matrix $\textbf{M} \in \{0,1\}^{N \times N_f}$ and a per-point face type matrix $\textbf{T} \in \{0,1\}^{N \times N_{f_t}}$, ($N_f$ denotes the number of faces and $N_{f_t}$ is the number of B-Rep face types). The above can be equivalently formulated as learning, for each $\textbf{w}_i$, the mapping $\Phi_{t_2}(\textbf{w}_i)~=~\{\Tilde{\textbf{M}}^p_{i,:}, \Tilde{\textbf{T}}^p_{i,:}\}$, where $\Tilde{\textbf{M}}^p_{i,:}~\in~[0,1]^{N_f}$ encodes the estimated probabilities of the $i$'th point belonging to one of the $N_f$ faces, and $\Tilde{\textbf{T}}^p_{i,:}~\in~[0,1]^{N_{f_t}}$ denotes the face type probabilities of the $i$'th point. This results in two probability matrices $\Tilde{\textbf{M}}^p~\in~[0,1]^{N \times N_f}$ and $\Tilde{\textbf{T}}^p~\in~[0,1]^{N \times N_{f_t}}$ for a scan $\textbf{X}$.

To learn $\Tilde{\textbf{T}}^p$, a standard categorical cross-entropy loss is employed \textit{w.r.t} the ground truth per-point face types. Learning $\Tilde{\textbf{M}}^p$ for face membership prediction additionally requires addressing the inherent ambiguity in face membership labelling, as face labels are arbitrary and do not necessarily have a semantic meaning. Inspired by~\cite{li2019supervised}, a Hungarian matching~\cite{hungarian} step is adopted to match the predicted face grouping to the ground truth by computing a Relaxed Intersection over Union (RIoU)~\cite{li2019supervised} cost between the predicted membership probabilities and the ground truth face labels. 
Following class assignment recovery, the RIoU is further employed in a loss function to learn the grouping as in~\cite{li2019supervised}. Finally, the estimated face memberships $\Tilde{\textbf{M}}~\in~\{0,1\}^{N \times N_f}$ and face types $\Tilde{\textbf{T}}~\in~\{0,1\}^{N \times N_{f_t}}$ are obtained through majority voting from $\Tilde{\textbf{M}}^p$ and $\Tilde{\textbf{T}}^p$, respectively. 
Note that since $\textbf{X}$ is downsampled with $N=10k$ during training, an upsampling of the predictions to the original scan resolution is used at inference.

\vspace{0.2cm}
\textbf{Track 3 Baseline.}
The goal of this track is to predict a point-to-step membership matrix $\textbf{S}~\in~\{0,1\}^{N \times N_s}$ and a per-point CAD operation matrix $\textbf{O}~\in~\{0,1\}^{N \times N_{O_t}}$, where $N_s$ denotes the number of the steps used in the CAD history and $N_{O_t}$ is the number of CAD operation types considered. Similarly to \textit{Track~2}, the above can be equivalently formulated as learning, for each per-point representation $\textbf{w}_i$, the mapping $\Phi_{t_3}(\textbf{w}_i) = \{ \Tilde{\textbf{S}}^p_{i,:}, \Tilde{\textbf{O}}^p_{i,:}\}$ where $\Tilde{\textbf{S}}^p_{i,:}~\in~[0,1]^{N_s}$ encode the estimated probabilities of the $i$'th point belonging to one of the $N_s$ CAD operation steps, and $\Tilde{\textbf{O}}^p_{i,:}~\in~[0,1]^{N_{O_t}}$ denotes the CAD operation type probabilities of the $i$'th point. This yields two probability matrices $\Tilde{\textbf{S}}^p~\in~[0,1]^{N \times N_s}$ and $\Tilde{\textbf{O}}^p~\in~[0,1]^{N \times N_{O_t}}$ for a scan $\textbf{X}$. Compared to the proposed solution for Section~\ref{track2_baseline}, the operation steps are ordered and thus, there is no need to address any labelling ambiguity. Hence, both $\Tilde{\textbf{S}}^p$ and $\Tilde{\textbf{O}}^p$ can be learned through a standard categorical cross-entropy loss. The final predicted operation steps $\Tilde{\textbf{S}}~\in~\{0,1\}^{N \times N_s}$ and types $\Tilde{\textbf{O}}~\in~\{0,1\}^{N \times N_{O_t}}$ are obtained by majority voting from the learned probability matrices $\Tilde{\textbf{S}}^p$ and $\Tilde{\textbf{O}}^p$, respectively. Note that the upsampling of the predictions is also performed here as described in Section~\ref{track2_baseline} for \textit{Track~2}. 


\vspace{-0.2cm}
\section{Evaluation Metrics}
\label{sect:metrics}
\vspace{-0.1cm}
To evaluate produced solutions, a set of dedicated metrics are considered to form a final score between $0$ and $1$. Evaluation is conducted in th Codalab platform\footnotemark[4]$^{,}$\footnotemark[5]$^{,}$\footnotemark[6].

\footnotetext[4]{\href{https://codalab.lisn.upsaclay.fr/competitions/13629}{https://codalab.lisn.upsaclay.fr/competitions/13629}
}

\footnotetext[5]{\href{https://codalab.lisn.upsaclay.fr/competitions/13956}{https://codalab.lisn.upsaclay.fr/competitions/13956}
}

\footnotetext[6]{\href{https://codalab.lisn.upsaclay.fr/competitions/13676}{https://codalab.lisn.upsaclay.fr/competitions/13676}
}

\vspace{0.2cm}
\textbf{Evaluation Metrics for Track 1}.
The evaluation of \textit{Track 1} consists of assessing the quality of the estimated parametric edges $\{\Tilde{\textbf{e}_{j}}\}_{j=1}^{\Tilde{N}_e}$ and their predicted sharpness with respect to the ground truth $\{{\textbf{e}_{j}}\}_{j=1}^{N_e}$. Note that the number of predicted edges $\Tilde{N}_e$ can be different from the ground truth number ${N}_e$. This is achieved following three criteria that are described in the following. For notation simplicity, we will denote the predicted set of edges by $\Tilde{\textbf{e}}$ and the ground truth ones by ${\textbf{e}}$. 



\textit{Edge Recovery Score:} This score measures the similarity between the predicted and the ground truth sets of parametric edges denoted by $\Tilde{\textbf{e}}$ and ${\textbf{e}}$, respectively. Given the parametric formulation of the predicted and ground truth edges described in Section~\ref{sec:track1}, the first step is to uniformly sample a set of 3D points on these edges proportionally to their length. This results in two 3D point sets $\Tilde{\mathrm{Z}} = \{\Tilde{{\boldsymbol{\zeta}}}_i\}_{i=1}^{\Tilde{N}_{\zeta}}$ and $\mathrm{Z} = \{{{\boldsymbol{\zeta}}}_i\}_{i=1}^{{N}_{\zeta}}$ for the predicted and ground truth edges, respectively. Then, two unidirectional Chamfer distances~\cite{achlioptas2018learning} are separately computed between the sampled point set on the predicted edges $\Tilde{\mathrm{Z}}$ and the sampled one on the ground truth  $\mathrm{Z}$ as follows, 
\vspace{-0.2cm}
\begin{equation}
d_{CD}\left(\Tilde{\mathrm{Z}}, \mathrm{Z}\right)=\frac{1}{\Tilde{N}_{\zeta}D_{\Tilde{\mathrm{Z}}}} \sum_{i=1}^{\Tilde{N}_{\zeta}} \min _{\boldsymbol{\zeta}_j \in \mathrm{Z}}\|\Tilde{{\boldsymbol{\zeta}}}_i-\boldsymbol{\zeta}_j\|_2^2 \ , 
    \label{eq:CD1}
\end{equation}
\vspace{-0.2cm}
\begin{equation}
    d_{CD}\left(\mathrm{Z}, \Tilde{\mathrm{Z}}\right)=\frac{1}{N_{\zeta}D_{\mathrm{Z}}} \sum_{i=1}^{N_{\zeta}} \min _{\boldsymbol{\Tilde{\zeta}}_j \in \Tilde{\mathrm{Z}}}\|{\boldsymbol{\zeta}}_i-\boldsymbol{\Tilde{\zeta}}_j\|_2^2 \ , 
    \label{eq:CD2}
\end{equation}

\noindent where $D_{\Tilde{\mathrm{Z}}}$ and $D_{\mathrm{Z}}$ denote the length of the diagonal of the bounding box of ${\Tilde{\mathrm{Z}}}$ and $\mathrm{Z}$, respectively. The obtained Chamfer distances further are normalized through a function $\Phi_{k}(d_{CD}) = e^{-kd_{CD}}$, that maps a distance $d_{CD}$ to a score in $[0,1]$, where the parameter $k$ is chosen according to conducted baselines. The two unidirectional scores are finally averaged to obtain a single edge recovery score 
\vspace{-0.2cm}
\begin{equation}
S_{e}(\Tilde{\textbf{e}}, \textbf{e}) =\frac{1}{2}(\Phi_{k}(d_{CD}(\Tilde{\mathrm{Z}},\mathrm{Z})) + 
\Phi_{k}(d_{CD}(\mathrm{Z},\Tilde{\mathrm{Z}}))) \ . 
\label{eq:edge-dist-track1}
\end{equation}

\noindent Note that accurately predicted edges close to the ground truth are expected to have a high edge recovery score $S_e$.




 




\textit{Edge length Score:} This score quantifies the similarity between the total edge length of the prediction $\Tilde{\textbf{e}}$ and that of the ground truth $\textbf{e}$. The length of the predicted and ground truth edges denoted as $\Tilde{L}$ and $L$, respectively, are computed by summing over the lengths of all edges in $\Tilde{\textbf{e}}$  and 
$\textbf{e}$. Note that for splines, the length is estimated by 
densely sampling points on the spline and then accumulating the 
consecutive point distances. The normalized length score is given by 
\vspace{-0.2cm}
\begin{equation}
S_l(\Tilde{\textbf{e}}, \textbf{e}) = 1 - |  \frac{1 -  (\Tilde{L} / L )}{1 +  (\Tilde{L} / L )}  | \ ,  
\label{eq:edge-length-track1}
\end{equation}

\noindent Note that this score has a range of $[0,1]$. Predicted edges with accurate length estimations are expected to have high length scores $S_l$. 


\textit{Sharpness Score:} The sharpness estimation task is formulated as a binary classification problem where each predicted edge $ \Tilde{\textbf{e}_{j}} \in \Tilde{\boldsymbol{e}}$ is classified as either sharp ($\Tilde{s}_j=1$) or not-sharp ($\Tilde{s}_j=0$). Since ground truth sharpness is given as a continuous value (ranging in $[0,2\pi]$), we use a threshold of $1.5$, above which an edge is considered sharp. The sharpness score is defined as a weighted accuracy of the predicted sharpness scores $\Tilde{\textbf{s}}$ with respect to the ground truth $\textbf{s}$. In practice, similarly to the edge recovery score, we sample two 3D point sets $\Tilde{\mathrm{Z}} = \{\Tilde{{\boldsymbol{\zeta}}}_i\}_{i=1}^{\Tilde{N}_{\zeta}}$ and $\mathrm{Z} = \{{{\boldsymbol{\zeta}}}_i\}_{i=1}^{{N}_{\zeta}}$ for the predicted and ground truth edges, respectively. Since the sharpness label is defined per edge, this label is transferred to the 3D points that form that edge. This results in a set of ground truth sharpness labels $\{\sigma_{i}\}_{i=1}^{i=N_{\zeta}} \in \{0,1\}$, where each $\sigma_{i}$ denotes the sharpness label of the point $\boldsymbol{\zeta}_i$. Similarly to the ground truth, another set of predicted sharpness labels $\{\Tilde{\sigma}_{i}\}_{i=1}^{i=\Tilde{N}_{\zeta}} \in \{0,1\}$ is produced from the predicted edge sharpness labels $\Tilde{\textbf{s}}$. The sharpness score is then given by the following weighted accuracy, 
%
\vspace{-0.2cm}
\begin{equation}
S_s = \frac{1}{\Tilde{N}_{\zeta}} \sum_{i=1}^{\Tilde{N}_{\zeta}} \Phi_{k}(\|\Tilde{{\boldsymbol{\zeta}}}_i-\boldsymbol{\zeta}_{\Gamma(i)}\|_2^2) \cdot 
\mathbbm{1}(\Tilde{\sigma}_{i} = \sigma_{\Gamma(i)}) \ ,     
\label{eq:sharpness-track1}
\end{equation}

\noindent where $\Gamma(i)$ matches an index $i$ of a point $\Tilde{{\boldsymbol{\zeta}}}_i$ in the predicted edges to the index of the closest point in the ground truth edges in the sense of Euclidean distance. $\Phi_k$ is the same mapping function used in the edge recovery score and $\mathbbm{1}(.)$ is an indicator function.



\textit{Final Score:} The final score of \textit{Track 1} is a combination of the three scores mentioned above. For each sample, it is computed as the average of the edge recovery score $S_e$, the length score $S_l$, and the sharpness Score $S_s$, or $S_{track 1}~=~\frac{1}{3}~(S_e + S_l + S_s)$.


\begin{table*}[!t]
\setlength{\belowcaptionskip}{-0.2cm}
\setlength{\tabcolsep}{3pt}
\begin{subtable}{.65\linewidth}
\centering
\begin{tabular}{ccccccccccccc}

   & \multicolumn{4}{c}{\textbf{Track 1}} & & \multicolumn{3}{c}{\textbf{Track 2}} & & \multicolumn{3}{c}{\textbf{Track 3}} \\
\cmidrule{2-13}
    & $S_{e}$ & $S_{l}$ & $S_{s}$ & $S_{track 1}$ & & $S_{m}$ & $S_{t}$ & $S_{track 2}$ & & $S_{st}$ & $S_{ot}$ & $S_{track 3}$ \\   
\cmidrule{2-5}
\cmidrule{7-9}
\cmidrule{11-13}
\textit{Baseline} & 0.38 & 0.18 & 0.46 & \textbf{0.34} & & 0.28 & 0.35 & \textbf{0.32} & & 0.29 & 0.39 & \textbf{0.34}\\
  \bottomrule
  \end{tabular}
\end{subtable}
\begin{subtable}{.32\linewidth}
 \hspace{0.5cm}
  \begin{tabular}{lcc}
& \textbf{Consistency} \\
\midrule
  Face Type (\textit{Track 2})& 0.79\\
  Operation Type (\textit{Track 3})& 0.90\\
  \bottomrule
  \end{tabular}
  \end{subtable}
  \caption{\textit{(left)} Performance of our proposed baselines on the dedicated test partition for all three tracks of the challenge. \textit{(right)} Consistency as the percentage of vertices sharing the same face membership / operation step also having the same type. As in \cite{dupont2022cadops} sub-operation types are grouped into a single type, for example, \textit{extrude.end} and \textit{extrude.side}, into a single \textit{extrude}.}
  \label{table:trackres}
\end{table*}

\vspace{0.2cm}
\textbf{Evaluation Metrics for Track 2}. The goal of Tracks 2 is to predict the B-Rep face membership and the types for each point of the scan. Accordingly, we evaluate the predictions following two criteria that are described below. 

\vspace{0.2cm}

\textit{Face Membership Score:} Given a scan $\textbf{X}$, the predicted face membership $\Tilde{\textbf{M}} \in \{0,1\}^{N \times \Tilde{N}_f}$ is evaluated with respect to the ground truth face membership $\textbf{M} \in \{0,1\}^{N \times N_f}$ using Intersection Over Union (IoU). Note that the number of predicted face memberships $\Tilde{N}_f$ can be different from the one of the ground truth $N_f$. Moreover, the evaluation of the face membership segmentation task requires addressing the inherent ambiguity in face membership labelling, as predicted face labels do not necessarily have a predefined match with ground truth class labels. To handle this issue, we use the Hungarian matching algorithm~\cite{hungarian} to perform optimal matching between predicted face memberships and ground truth face memberships. Hungarian matching is able to find the best one-to-one correspondence that maximizes the total IoU across all matched pairs. This results in the following face membership score, 
\vspace{-0.2cm}
\begin{equation}
S_m = \frac{1}{N_f} \sum_{i=1}^{N_f} IoU(\Lambda(\Tilde{\textbf{M})}_{:,i},\textbf{M}_{:,i})  \ ,
\label{eq:membership-track2}
\end{equation}

\noindent where $\Lambda()$ is the reordering from Hungarian matching.

\textit{Face Type Score:} Similarly to the face membership score, the predicted face types $\Tilde{\textbf{T}}~\in~\{0,1\}^{N \times N_{f_t}}$ is evaluated with respect to the ground truth face types $\textbf{T}~\in~\{0,1\}^{N \times N_{f_t}}$ using IOU. However, in contrast to the face membership scenario where a Hungarian matching was necessary, the IoU is directly computed for the face types to yield the following face type score, 
\vspace{-0.2cm}
\begin{equation}
S_t = \frac{1}{N_{f_t}} \sum_{i=1}^{N_{f_t}} IoU(\Tilde{\textbf{T}}_{:,i},\textbf{T}_{:,i})  \ .
\label{eq:type-track2}
\end{equation}

\textit{Final Score:} The final score for each sample is the average of the face membership score $S_m$ and the face type score $S_t$ and is given by $S_{track 2} = \frac{1}{2} (S_m + S_t)$.
%









\vspace{0.2cm}
\textbf{Evaluation Metrics for Track 3}. As in \textit{Track 2}, the predicted CAD operation steps and types are evaluated following two criteria. 


\textit{CAD Step Score:} Given a scan $\textbf{X}$, the predicted face membership $\Tilde{\textbf{S}} \in \{0,1\}^{N \times \Tilde{N}_s}$ is evaluated with respect to the ground truth face membership $\textbf{S} \in \{0,1\}^{N \times N_s}$ using IoU. As the objective is to predict ordered steps, Hungarian matching is not used and the CAD step score is given by
\vspace{-0.2cm}
\begin{equation}
S_{st} = \frac{1}{N_{s}} \sum_{i=1}^{N_{s}} IoU(\Tilde{\textbf{S}}_{:,i},\textbf{S}_{:,i})  \ .
\label{eq:step-track3}
\end{equation}


\textit{CAD Type Score:} As done for the face types of Track~2, the predicted CAD operation types $\Tilde{\textbf{O}}~\in~\{0,1\}^{N \times N_{O_t}}$ is evaluated with respect to the ground truth face types $\textbf{O}~\in~\{0,1\}^{N \times N_{O_t}}$ using the following IoU based score
\vspace{-0.2cm}
\begin{equation}
S_{ot} = \frac{1}{N_{t}} \sum_{i=1}^{N_{t}} IoU(\Tilde{\textbf{O}}_{:,i},\textbf{O}_{:,i})  \ .
\label{eq:type-track3}
\end{equation}


\textit{Final Score:} The final score for each sample is the average of the CAD step score $S_{st}$ and the CAD type score $S_{ot}$ and is given by $S_{track 3} = \frac{1}{2} (S_{st} + S_{ot})$.
%






\begin{figure}[!ht]
\setlength{\belowcaptionskip}{-0.4cm}
\begin{center}
   \includegraphics[width=1.0\linewidth]{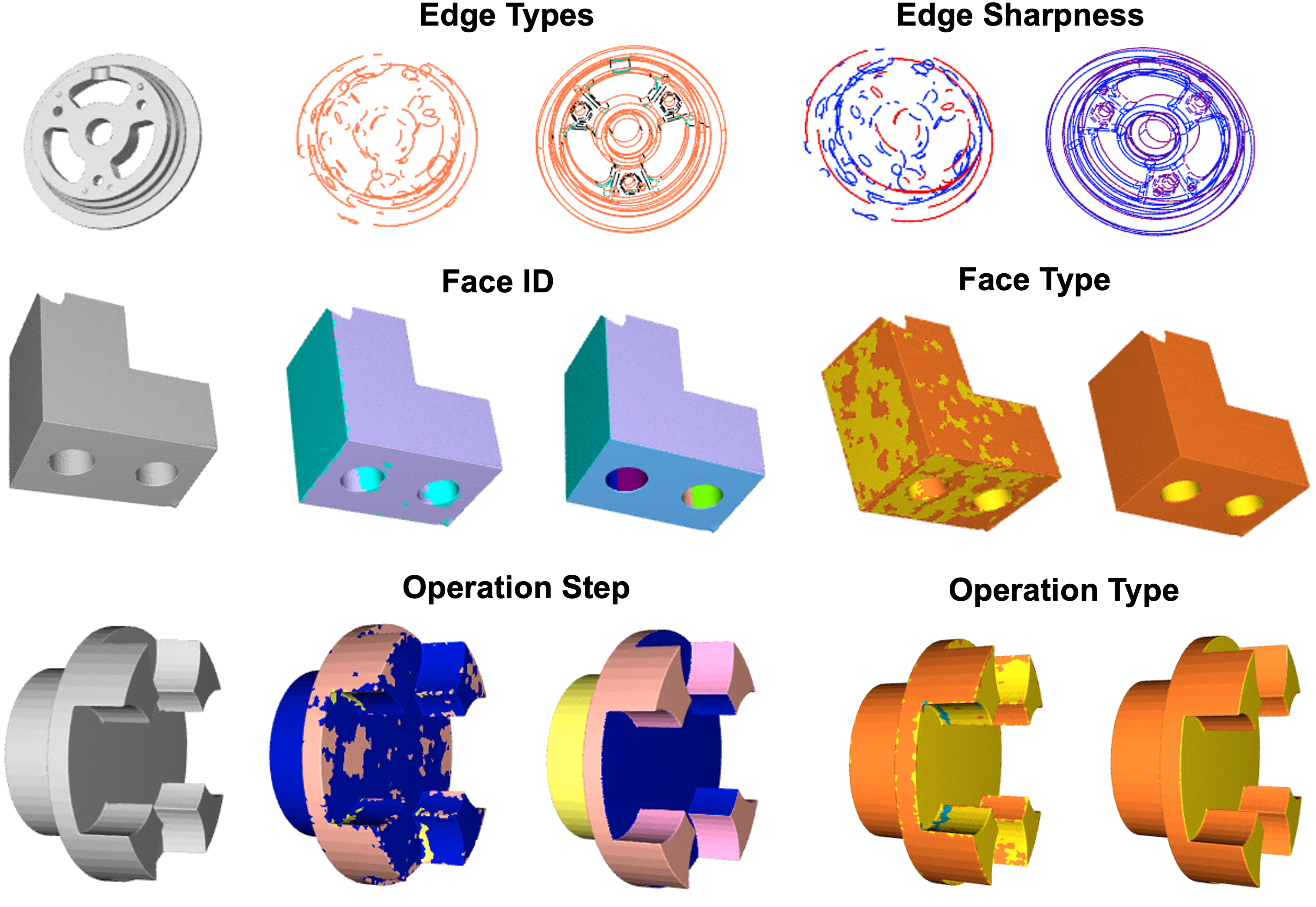}
\end{center}
\vspace{-0.1cm}
\caption{Qualitative results for proposed baselines for the three tracks of the challenge (one row per track). Model prediction (\textit{left}) is contrasted to the ground truth labels (\textit{right}). Colour labelling as in Figure \ref{fig:track_description}.}
\label{fig:visualres}
\end{figure}

\section{Results}
\label{sect:results}

The proposed baseline methods for all three tracks are evaluated on the dedicated test partition. Results are shown in Table \ref{table:trackres} (\textit{left}) and are also given on the online challenge leaderboard (hosted on the Codalab platform\footnotemark[4]$^{,}$\footnotemark[5]$^{,}$\footnotemark[6]). It is essential to note that the performance of the evaluated baselines is not particularly robust, with a reported final score of $0.34$ for \textit{Track 1}, and \textit{Track 3} and $0.32$ for \textit{Track 2}; We highlight that the primary motivation of this experimental analysis is to establish a reference point on baseline performance for all tracks, rather than striving to set new track records. Consequently, the baselines were not subject to optimization. For instance, we did not specifically address issues such as class imbalance, undertake extensive hyperparameter tuning or utilise the adaptive sampling scheme of \cite{Cherenkova2023SepicNetSE} for enhanced edge detection. For qualitative results on Figure~\ref{fig:visualres}, we observe that our baseline model tends to segment larger circles into fragmented sets of shorter edges, cluster face IDs together and often conflates distinct operation steps. These findings underscore the extent of the difficulty of the challenge and highlight potential areas for improvements in future iterations.

\begin{figure}
\setlength{\belowcaptionskip}{-0.5cm}
\begin{center}
   \includegraphics[width=.65\linewidth]{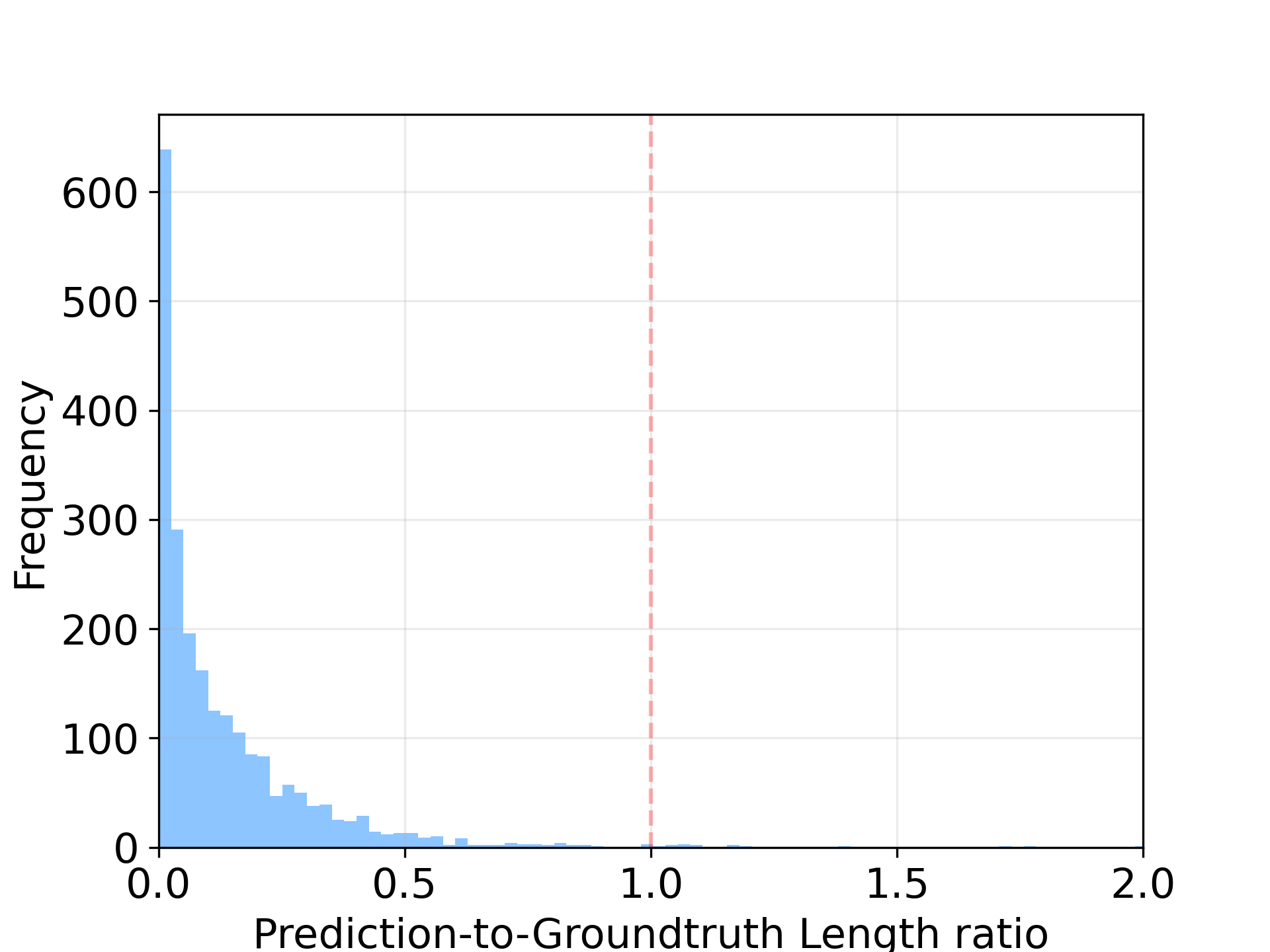}
\end{center}
\vspace{-0.6cm}
\caption{ Histogram of \textit{Prediction-to-Groundtruth} length ratios $\Tilde{L} / L$ across test samples.
}
\label{fig:lengthhist}
\end{figure}

\begin{figure}[!ht]
\setlength{\abovecaptionskip}{-0.2cm}
\setlength{\belowcaptionskip}{-0.5cm}
    \centering
    \begin{subfigure}{.49\linewidth}
        \centering
        \includegraphics[valign=t, width=0.8\linewidth]{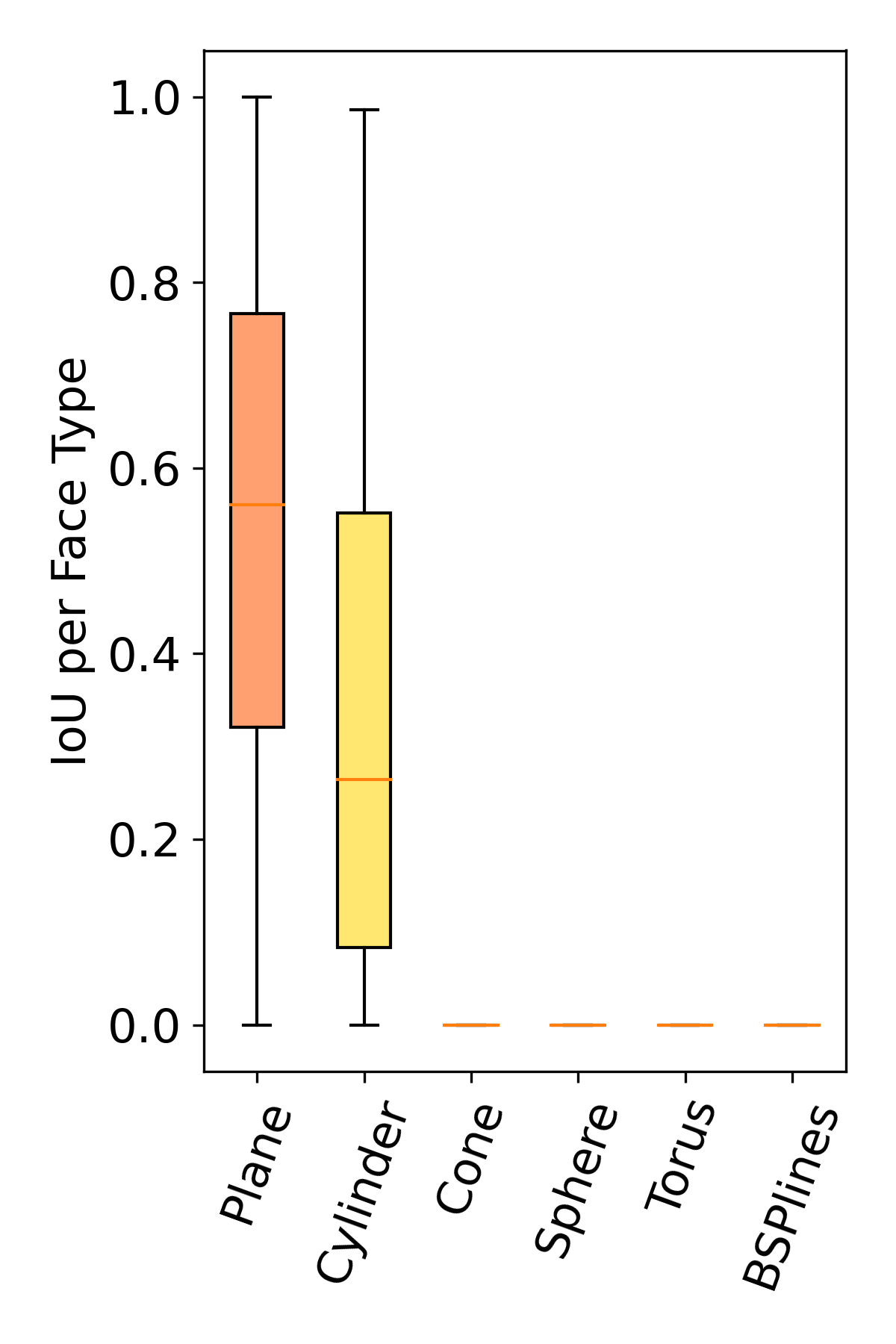}
    \end{subfigure}
    \begin{subfigure}{.49\linewidth}
        \centering
        \includegraphics[valign=t, width=0.85\linewidth]{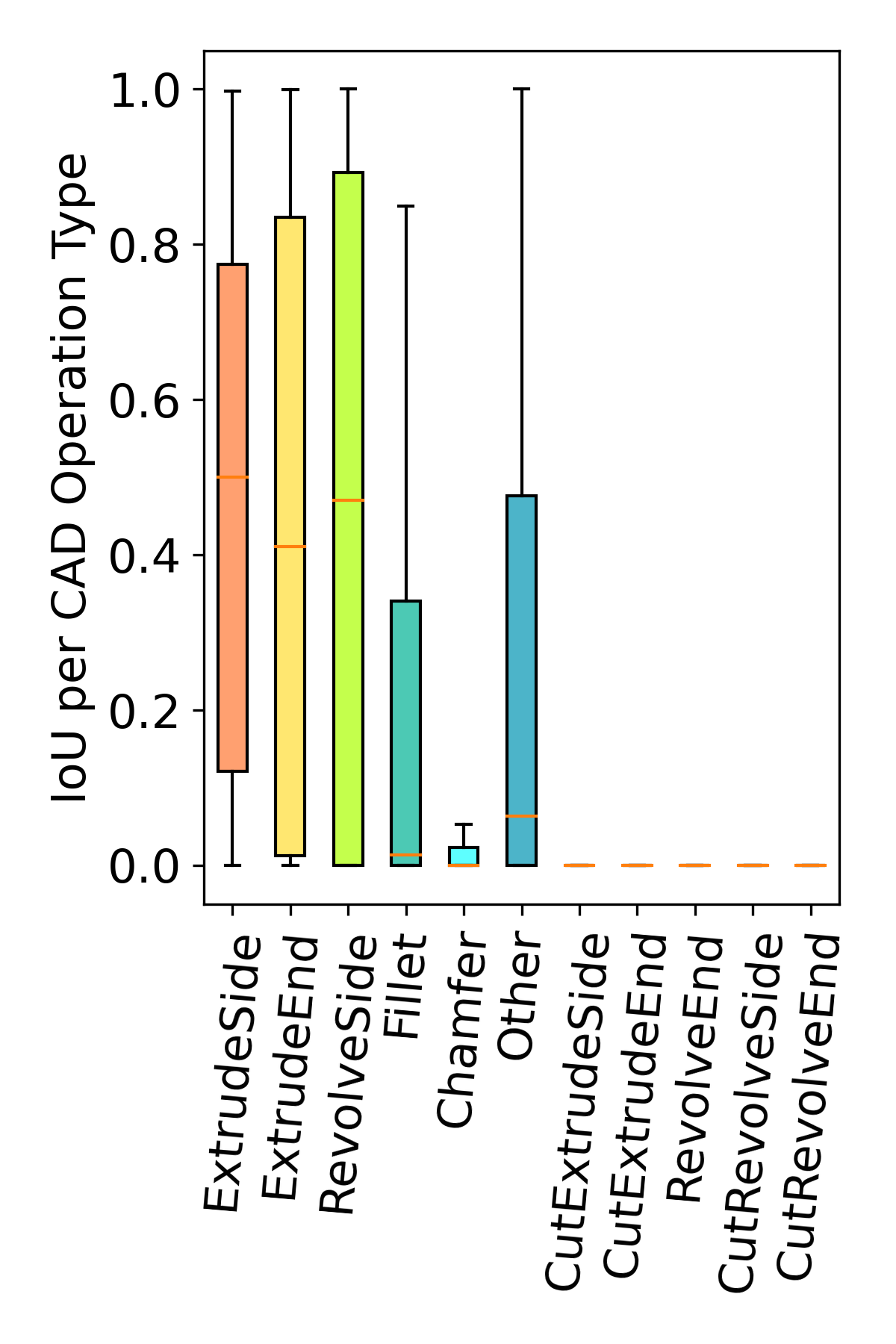}
    \end{subfigure}
    \caption{Intersection over Union IoU reported per type, for face types (\textit{Track 2}) and operation types (\textit{Track 3}).}
    \label{fig:typeperformance}
\end{figure}

To provide additional insights on model performance, we present a histogram of \textit{prediction-to-groundtruth} length ratios in Figure~\ref{fig:lengthhist}. We find that our baseline consistently underestimates edge lengths thus limiting performance. In Figure~\ref{fig:typeperformance}, we report per-type IoU for face types (\textit{Track 2}) and operation types (\textit{Track 3}). Our baseline struggles to capture less common types in both cases due to a significant imbalance in class frequency (as also shown in Figure~\ref{fig:track 2 stat graphs} and Figure~\ref{fig:track 3 stat graphs}). Finally, we follow \cite{dupont2022cadops} and report face and step prediction consistency in Table~\ref{table:trackres} \textit{(right)} as the percentage of vertices sharing the same face membership / operation step also having the same type. We identify that future improvements in terms of consistency (we report $0.79$ for \textit{Track~2} and $0.90$ for \textit{Track~3}) can positively affect performance.

\section{Conclusion}\label{sec:conclusions}

In this paper, we introduce the SHARP challenge 2023, aiming to address the nuances of the Scan-to-CAD problem through three distinct tracks. For every track, a new version of the challenging CC3D dataset is presented, along with an exhaustive description of the evaluation metrics and proposed baseline methodologies. This challenge is designed to encourage forthcoming advancements in reverse engineering from 3D scans in a real-world setting.

\vspace{0.2cm}
\textbf{Acknowledgement:} Present work is supported by the National Research Fund, Luxembourg under the BRIDGES2021/IS/16849599/FREE-3D and
IF/17052459/CASCADES projects, and by Artec 3D.

{\small
\bibliographystyle{ieee_fullname}
\bibliography{egbib}
}

\end{document}